\documentclass[11pt]{article}

\usepackage[preprint]{acl}

\usepackage{times}
\usepackage{latexsym}
\usepackage{enumitem}
\usepackage{booktabs}
\usepackage{array} 
\newcolumntype{Y}{>{\centering\arraybackslash}X}
\usepackage{tabularx} % 需要在导言区加这个包
\usepackage[T1]{fontenc}
\usepackage[utf8]{inputenc}

\usepackage{microtype}
\usepackage{multirow}

\usepackage{inconsolata}

\usepackage{graphicx}
\usepackage{booktabs}
\usepackage{amsmath,amssymb,amsfonts}
\usepackage{makecell}

\title{Consolidation or Adaptation? PRISM: Disentangling SFT and RL Data via Gradient Concentration}

\author{Yang Zhao$^{\spadesuit}\footnotemark[1]$,Yangou Ouyang$^{\spadesuit}\footnotemark[1]$, Xiao Ding$^{\spadesuit}\footnotemark[2]$,Hepeng Wang$^{\spadesuit}$,Bibo Cai$^{\spadesuit}$, Kai Xiong$^{\spadesuit}$,\\
Jinglong Gao$^{\spadesuit}$, Zhouhao Sun$^{\spadesuit}$, Li Du$^{\heartsuit}$, Bing Qin$^{\spadesuit}$ and  Ting Liu$^{\spadesuit}$ \\
  $^{\spadesuit}$Research Center for Social Computing and Interactive Robotics,\\ %
  Harbin Institute of Technology, China \\
  $^{\heartsuit}$Beijing Academy of Artificial Intelligence, Beijing, China\\
 \texttt{\{yangzhao, yooy\}@ir.hit.edu.cn}\\
 \\\\}

\begin{document}
\maketitle
\renewcommand*{\thefootnote}{\fnsymbol{footnote}}
\footnotetext[1]{These authors contributed equally to this work.}
\renewcommand*{\thefootnote}
{\fnsymbol{footnote}}
\footnotetext[2]{Corresponding Author.}
\renewcommand*{\thefootnote}
{\fnsymbol{footnote}}
\renewcommand*{\thefootnote}
{\arabic{footnote}}
\begin{abstract}

While Hybrid Supervised Fine-Tuning (SFT) followed by Reinforcement Learning (RL) has become the standard paradigm for training LLM agents, effective mechanisms for data allocation between these stages remain largely underexplored. Current data arbitration strategies often rely on surface-level heuristics that fail to diagnose intrinsic learning needs. Since SFT targets pattern consolidation through imitation while RL drives structural adaptation via exploration, misaligning data with these functional roles causes severe optimization interference. We propose PRISM, a dynamics-aware framework grounded in Schema Theory that arbitrates data based on its degree of cognitive conflict with the model's existing knowledge. By analyzing the spatial geometric structure of gradients, PRISM identifies data triggering high spatial concentration as high-conflict signals that require RL for structural restructuring. In contrast, data yielding diffuse updates is routed to SFT for efficient consolidation. Extensive experiments on WebShop and ALFWorld demonstrate that PRISM achieves a Pareto improvement, outperforming state-of-the-art hybrid methods while reducing computational costs by up to 3.22$\times$. Our findings suggest that disentangling data based on internal optimization regimes is crucial for scalable and robust agent alignment. The source code for this work is publicly available at: https://github.com/zy125413/PRISM.
\end{abstract}

\begin{figure}[t]
    \centering
    % 调整 width 参数以适应你的栏宽，通常 1.0\linewidth 或 0.95\linewidth
    \includegraphics[width=1.0\linewidth]{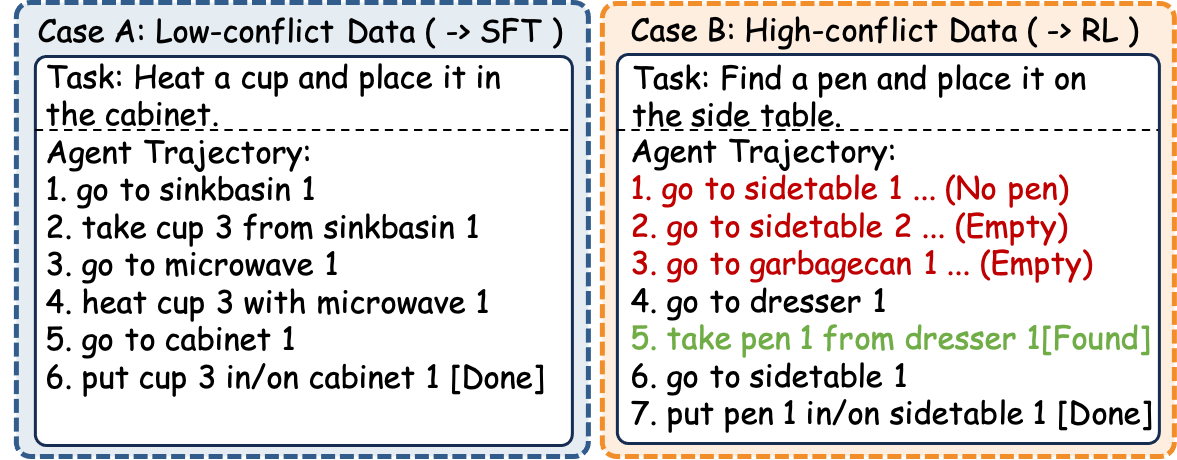}
\caption{Case Study on ALFWorld. PRISM performs data arbitration by diagnosing cognitive conflict between task trajectories and the model's internal state. (Left) Case A: A routine task follows a linear execution sequence, characterized by diffuse gradient updates (low concentration). Such samples are routed to SFT for behavioral consolidation. (Right) Case B: A high-conflict task involving extensive trial-and-error (e.g., searching multiple locations) triggers concentrated gradient updates (high concentration). These signals indicate a failure in the model's current logic, necessitating RL for structural adaptation and reasoning refinement. }
    \label{fig:case_study}
\end{figure}

\section{Introduction}
Large Language Model (LLM) agents have demonstrated remarkable capabilities in complex decision-making tasks~\citep{qian2025toolrl,qin2024toolllm}. To unlock these potentials, the prevailing paradigm has adopted a standard two-stage training pipeline: SFT to establish behavioral norms, followed by RL to optimize long-horizon exploration~\citep{zhang2025generalizability}. This pipeline relies on a functional synergy: SFT facilitates pattern consolidation by internalizing behavioral norms and task-specific knowledge, while RL drives structural adaptation via trial-and-error to refine logic and enhance generalization~\citep{chu2025sft}. 
Given the divergent optimization mechanisms of SFT (imitation) and RL (exploration), the alignment efficiency and effectiveness hinges on precisely partitioning data between these regimes according to their intrinsic cognitive demands~\citep{lv2025towards}.

However, efficiently partitioning data between SFT and RL remains a non-trivial challenge, as current paradigms are often constrained by three primary limitations:
(1) Monolithic Sequencing applies a fixed SFT-then-RL schedule to large instruction corpora~\citep{ouyang2022training}. This uniform approach ignores data heterogeneity, leading to computational inefficiencies by failing to distinguish between data requiring behavioral consolidation and that necessitating exploratory reasoning~\citep{zhou2023lima}. (2) Universal Exploration~\citep{shao2024deepseekmath, feng2025group} subjects broad trajectories to RL indiscriminately. Yet, applying trial-and-error to high-conflict data without SFT-consolidated behavioral priors can trigger optimization instability, hindering the formation of coherent reasoning pathways~\citep{deepseek2025r1}. (3) Outcome-Centric Filtering~\citep{lv2025towards} relies on external proxies (e.g., accuracy) to estimate knowledge conflict. This creates an observational gap where external correctness masks latent shortcut learning, where the model attains answers via spurious cues rather than through faithful reasoning~\citep{geirhos2020shortcut, dziri2023faith}. Consequently, these proxies fail to capture true model--data conflict, overlooking examples that require genuine structural adaptation~\citep{dai2022knowledge}.
Across these regimes, the fundamental bottleneck is that data routing relies on coarse heuristics, such as pipeline order or outcome-based indicators, rather than intrinsic signals reflecting the model’s internal state.

Inspired by Schema Theory~\citep{piaget1952origins}, we posit that learning efficiency is fundamentally governed by the degree of conflict between new information and the model’s existing knowledge base. In this framework, compatible information is mastered through consolidation, while high-conflict information necessitates a fundamental restructuring of internal logic.
To operationalize this principle, we adopt a gradient-based perspective, viewing gradients as the mathematical "feedback" signal derived from data. We propose that the distributional geometry of gradients serves as a critical proxy for this cognitive conflict. Specifically, we utilize statistical metrics such as the Gini coefficient to quantify gradient concentration. Prior studies suggest that concentrated updates (high Gini) typically correspond to data that deviates significantly from the model's established knowledge base~\citep{simsekli2019tail}, whereas diffuse updates reflect global consistency with its current parametric state~\citep{chizat2019lazy}.
As illustrated in our ALFWorld case study (Figure \ref{fig:case_study}), this gradient signal enables precise Data Arbitration. A "low-conflict" task (Case A) follows a standard routine and triggers diffuse gradients, making it an ideal candidate for SFT to consolidate behavioral norms. In contrast, a "high-conflict" scenario (Case B) involving complex error recovery generates highly concentrated gradients, signaling a failure of current logic that demands RL-driven exploration. By routing samples based on these intrinsic learning needs, \textbf{PRISM} (\textbf{P}artitioning \textbf{R}egimes via \textbf{I}nternal \textbf{S}patial-gradient \textbf{M}etrics) ensures both training efficiency and robust generalization.

We evaluate PRISM on two challenging agent benchmarks: WebShop (online shopping)~\citep{yao2022webshop} and ALFWorld (embodied decision-making)~\citep{shridhar2020alfworld}.
Empirical results demonstrate that PRISM achieves a Pareto improvement: it establishes a new state-of-the-art on ALFWorld (95.31) while reducing RL computational overhead by up to 3.22$\times$.
These results confirm that selective structural adaptation is both more robust and more efficient than exhaustive exploration.
Notably, PRISM exhibits superior generalization capabilities across diverse backbones, including Qwen and Llama architectures. By precisely allocating high-conflict data to RL, the model achieves substantial performance gains in unseen environments, strongly validating the importance of distinguishing between consolidation and adaptation data for building robust agents.

Our contributions are summarized as follows: 
\begin{itemize} [nosep,leftmargin=*] 
\item \textbf{Misalignment:} Formalizing the agent training bottleneck as functional SFT-RL mismatch from coarse-grained data allocation.
\item \textbf{PRISM Framework:} A framework using spatial geometric structure of gradients to diagnose cognitive conflict, routing data between consolidation and adaptation regimes.
\item \textbf{Efficiency:} We demonstrate SOTA performance across diverse backbones while delivering a \textbf{3.22$\times$} training speedup via selective high-conflict allocation. \end{itemize}

\begin{figure*}[t]
    \small
    \centering
    \includegraphics[width=1\linewidth]{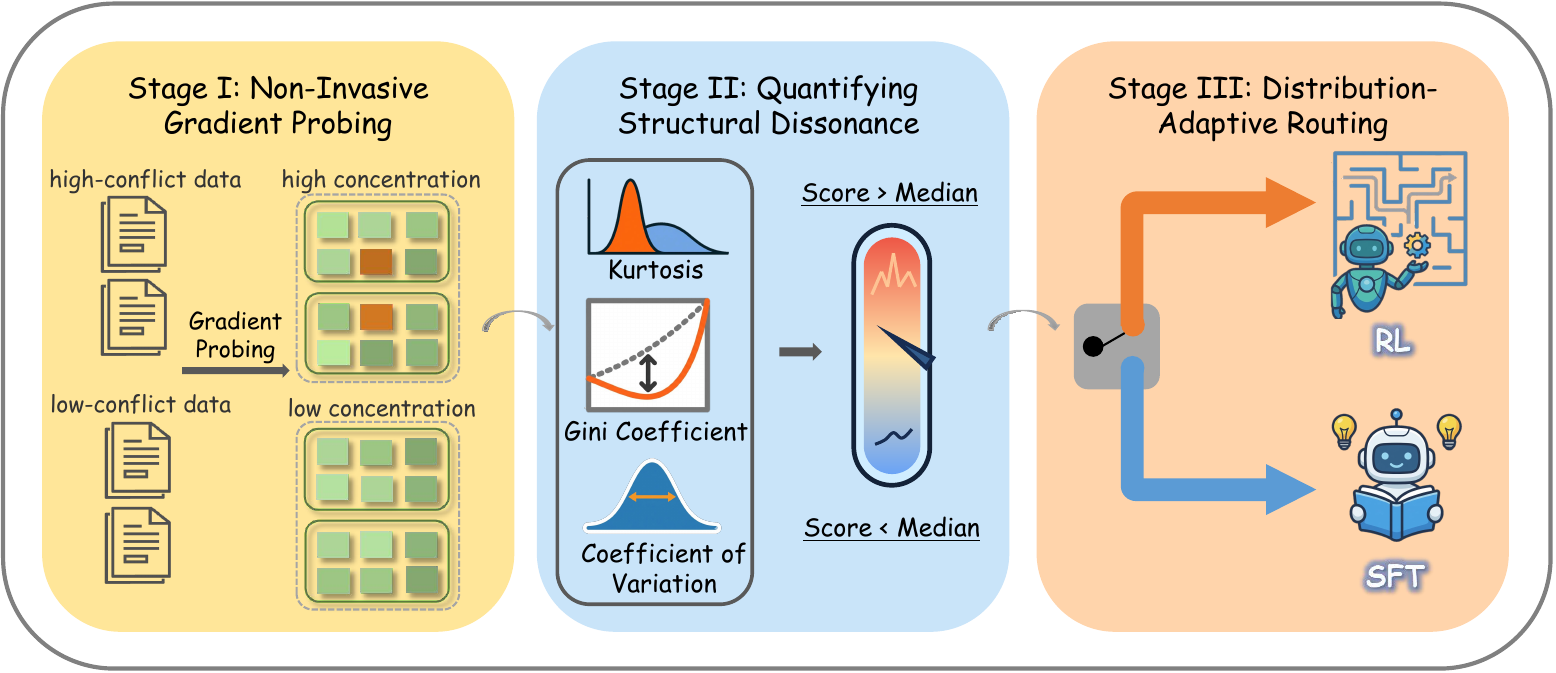}
    \caption{\textbf{Overview of PRISM.} The framework consists of three stages: (1) Non-Invasive Gradient Probing: Extracting update landscapes to capture internal reactions to each sample; (2) Quantifying Structural Dissonance: Measuring gradient concentration to diagnose the  conflict between the data and existing knowledge; (3) Distribution-Adaptive Routing: Partitioning data based on concentration—samples with low-conflict (diffuse updates) are routed to SFT for consolidation, while those with high conflict (concentrated updates) are routed to RL for structural restructuring.} 
    \label{fig:main_framework}
\end{figure*}

\section{Methodology}
\label{sec:method}

We formalize PRISM, a framework designed to operationalize data routing by distinguishing between pattern consolidation and structural adaptation. We first establish the theoretical foundation by defining \textbf{gradient concentration} as a diagnostic proxy for \textbf{cognitive conflict}. Building upon this groundwork, we detail the implementation of PRISM via a three-stage pipeline (Fig.~\ref{fig:main_framework}): (i) non-invasive gradient probing to capture the spatial geometric structure of learning dynamics; (ii) quantifying structural dissonance via statistical concentration metrics; and (iii) distribution-adaptive routing to allocate trajectories to their optimal learning regimes.

\subsection{Gradient Concentration as a Proxy for Cognitive Conflict}

This framework bridges the gap between low-level optimization dynamics and high-level cognitive learning processes by treating the spatial geometric structure of gradients as a diagnostic signal for the conflict between input data and the model’s existing knowledge schema. This logic is rooted in the functional specificity of model parameters: research indicates that knowledge representation in Transformers is often localized within a sparse subset of specific ``knowledge neurons''~\citep{dai2022knowledge}. When new information contradicts established patterns, the optimization process forces gradients into a significantly non-uniform spatial distribution, concentrating heavily within specific critical units to resolve internal logical dissonance~\citep{simsekli2019tail, meng2022locating}. Consequently, this spatial concentration serves as an effective proxy for the magnitude of internal structural adaptation required by the model.

\textbf{High Gradient Concentration} signifies that the model must undergo intense localized updates to reconcile fundamental contradictions between the input and existing heuristics, signaling a regime of Structural Restructuring. In these high-conflict scenarios, Reinforcement Learning (RL) is indispensable, as its exploration mechanism drives the deep policy shifts necessary for logic realignment. Conversely, \textbf{Low Gradient Concentration} corresponds to diffuse, low-intensity parameter updates across the network, implying that new information can be seamlessly integrated without overhauling the underlying knowledge architecture~\citep{chizat2019lazy}. This state represents knowledge compatibility and pattern consolidation, where Supervised Fine-Tuning (SFT) provides an efficient and stable optimization path to refine behavioral norms at a lower computational cost and with minimal risk of destructive interference. By utilizing concentration metrics to distinguish between these internal regimes, PRISM enables principled \textbf{Data Arbitration}, selectively deploying RL only when structural adaptation is essential.

\subsection{Stage I: Non-Invasive Gradient Probing}
To capture the model's internal reaction to new data without altering its weights, we perform Non-Invasive Gradient Probing. This stage serves as a lightweight ``diagnostic scan'' of the gradient landscape to identify the required optimization effort for each interaction trajectory $\tau_i$. 

Specifically, we utilize the ground-truth reference trajectories provided by the dataset as the supervisory signal. The rationale is to measure the dissonance between the model's current policy and the expert behavior required by the task. We decompose the model's parameter space $\Theta$ into $N$ fine-grained functional units. For a Transformer with $L$ layers, we focus on the linear projection matrices within the Attention and FFN blocks (e.g., $W_q, W_k, W_v, W_o, W_{\text{gate}}, \dots$), resulting in $N = 7L$ distinct parameter groups. 

To eliminate confounds arising from varying sequence lengths, all input sequences are processed using a uniform, task-specific context length. We apply strict attention masking to ensure that gradients are computed solely on valid response tokens, excluding padding artifacts. We execute a single backward pass without performing any optimizer update, and then aggregate matrix-wise Frobenius norms to obtain a high-fidelity ``snapshot'' of the internal learning state. This yields a high-dimensional gradient vector $\mathbf{g}_i \in \mathbb{R}_{\ge 0}^N$:
\begin{equation}
 \mathbf{g}_i = [ \| \nabla_{\theta_1} \mathcal{L}(\tau_i) \|_F, \dots, \| \nabla_{\theta_N} \mathcal{L}(\tau_i) \|_F ]^\top,
 \end{equation}
where $\mathcal{L}(\cdot)$ denotes the next-token prediction loss averaged over the valid response tokens, and $\| \nabla_{\theta_j} \mathcal{L}(\tau_i) \|_F$ is the Frobenius norm of the gradient with respect to the $j$-th functional group $\theta_j$. Significantly, this probing phase is computationally efficient and memory-friendly. By calculating the norms on-the-fly during the backward pass and discarding the full gradient tensors, we maintain a memory footprint nearly identical to a standard forward pass, resulting in negligible overhead: the probing step accounts for only about 1--2\% of the end-to-end wall-clock time of our full pipeline (probe+SFT+RL), as reported in Table~\ref{tab:raw_time_stats}.

\subsection{Stage II: Quantifying Structural Dissonance}

While the raw gradient vector $\mathbf{g}_i$ encodes both update intensity and structural shape, we prioritize the distributional shape to reveal the nature of the learning conflict, adhering to the principle of scale invariance. To robustly quantify this signal, we employ a statistical concentration toolkit comprising three complementary metrics. The Gini Coefficient measures the global inequality of gradient contributions across the network. Simultaneously, Kurtosis serves as a high-order diagnostic tool to detect heavy-tailed updates, identifying trajectories that force spiky adjustments in localized knowledge neurons while leaving the majority of functional circuits dormant. Finally, the Coefficient of Variation (CV) captures relative instability by normalizing dispersion against the mean update intensity. Together, these metrics triangulate cognitive dissonance from distinct statistical dimensions: high values signal an acute structural conflict requiring exploratory restructuring via RL, while low values indicate high data-model compatibility suitable for efficient consolidation via SFT.

In practice, for each trajectory $\tau_i$, we compute a composite score $s_i = \phi(\mathbf{g}_i)$, where $\phi(\cdot)$ is a statistical concentration operator (e.g., Gini) that maps the high-dimensional spatial geometric structure to a scalar value of cognitive dissonance.

\subsection{Stage III: Distribution-Adaptive Routing}

Finally, we partition the full corpus $\mathcal{D}$ into disjoint subsets based on the quantified dissonance using a distribution-adaptive strategy. We compute the global statistics of the concentration scores $\mathcal{S}$ and employ a non-parametric median split to define the routing boundary:
\begin{align}
\mathcal{D}_{\text{SFT}} &= \{ \tau_i \in \mathcal{D} \mid s_i \le \text{Median}(\mathcal{S}) \}, \\
\mathcal{D}_{\text{RL}}  &= \{ \tau_i \in \mathcal{D} \mid s_i > \text{Median}(\mathcal{S}) \},
\end{align}
where $\mathcal{D}$ is the initial training corpus, $\mathcal{S} = \{s_1, \dots, s_{|\mathcal{D}|}\}$ represents the global set of composite concentration scores for all trajectories, $s_i$ is the structural dissonance score for the $i$-th trajectory, and $\text{Median}(\mathcal{S})$ serves as the threshold that dynamically partitions the corpus into consolidation ($\mathcal{D}_{\text{SFT}}$) and adaptation ($\mathcal{D}_{\text{RL}}$) regimes.

We employ a non-parametric median split as a robust, data-adaptive thresholding strategy. This approach ensures that data arbitration is grounded in the relative cognitive dissonance of the specific corpus, eliminating the need for per-task hyperparameter tuning while maintaining a stable balance between plasticity and stability.

The rationale for selecting the median as the decision boundary is two-fold. First, this non-parametric boundary adaptively scales with the dataset's inherent difficulty, ensuring that arbitration is determined by the relative cognitive effort required by the model's current state avoiding arbitrary constants. Second, it strikes a theoretical balance between stability and plasticity: routing too many samples to RL (a low threshold) introduces optimization noise from easy data, while routing too few (a high threshold) limits capacity for structural adaptation. This design choice is empirically validated in our Ablation Studies (Section \ref{sec:ablation_ratio}), where the median split consistently yields optimal performance compared to extreme allocation ratios. Consequently, high-conflict trajectories are routed to RL for Structural Restructuring, while low-conflict trajectories are assigned to SFT for pattern consolidation, ensuring that each learning regime addresses the data's intrinsic conflict.

\begin{table*}[t]
\centering
\small
\renewcommand{\arraystretch}{1.2} % 进一步微调行高，提升阅读舒适度
\setlength{\tabcolsep}{0pt}

\begin{tabular*}{\textwidth}{@{\extracolsep{\fill}} llc ccccccc}
\toprule
& & \textbf{Split (\%)} & \multicolumn{7}{c}{\textbf{Task-wise Success Rate (\%)}} \\
\cmidrule{3-3} \cmidrule{4-10}
\textbf{Category} & \textbf{Method} & \textbf{SFT : RL} & \textbf{Pick} & \textbf{Look} & \textbf{Clean} & \textbf{Heat} & \textbf{Cool} & \textbf{Pick2} & \textbf{All} \\
\midrule
\multicolumn{10}{c}{\textit{(a) ALFWorld - Seen (In-Distribution)}} \\
\midrule
Base & Base Model & - : - & 39.29 & 7.14 & 0.00 & 0.00 & 0.00 & 0.00 & 9.38 \\
\midrule
Single Phase & SFT & 100 : - & 78.57 & 78.57 & 42.86 & 16.67 & 26.92 & 28.57 & 45.31 \\
& GRPO & - : 100 & 80.65 & 28.57 & 86.21 & 52.94 & 66.67 & 57.89 & 67.19 \\
& GiGPO & - : 100 & 87.10 & 42.86 & 75.86 & 76.47 & 61.11 & 52.63 & 69.53 \\
\midrule
Hybrid & Random & 50 : 50 & 92.59 & 66.67 & 80.00 & 61.54 & 75.76 & 88.00 & 79.69 \\
& HPT & 50 : 50 & 92.59 & 77.78 & 92.59 & 75.00 & 75.00 & 75.00 & 85.16 \\
& SFT-then-RL & 100 : 100 & 96.67 & 72.73 & 96.15 & 85.71 & 80.00 & 86.36 & 88.28 \\
\midrule
\textbf{PRISM} & Gini & 50 : 50 & 92.68 & 100.00 & 100.00 & 100.00 & 95.00 & 91.67 & \textbf{95.31} \\
\textbf{(Ours)} & Kurtosis & 50 : 50 & 95.12 & 88.89 & 100.00 & 100.00 & 75.00 & 87.50 & 91.41 \\
& CV & 50 : 50 & 95.12 & 88.89 & 100.00 & 90.00 & 85.00 & 75.00 & 89.84 \\

\midrule[1pt]
\multicolumn{10}{c}{\textit{(b) ALFWorld - Unseen (Out-of-Distribution)}} \\
\midrule
Base & Base Model & - : - & 15.00 & 13.33 & 2.56 & 0.00 & 0.00 & 0.00 & 4.69 \\
\midrule
Single Phase & SFT & 100 : - & 70.00 & 20.00 & 15.38 & 4.17 & 10.00 & 0.00 & 19.53 \\
& GRPO & - : 100 & 75.00 & 53.33 & 92.31 & 75.00 & 60.00 & 20.00 & 67.97 \\
& GiGPO & - : 100 & 90.00 & 53.33 & 76.92 & 75.00 & 60.00 & 40.00 & 68.75 \\
\midrule
Hybrid & Random & 50 : 50 & 45.00 & 33.33 & 74.36 & 54.17 & 90.00 & 65.00 & 60.94 \\
& HPT & 50 : 50 & 80.00 & 46.67 & 84.62 & 66.67 & 80.00 & 70.00 & 73.44 \\
& SFT-then-RL & 100 : 100 & 65.00 & 60.00 & 71.79 & 58.33 & 100.00 & 75.00 & 69.53 \\
\midrule
\textbf{PRISM} & Gini & 50 : 50 & 75.00 & 60.00 & 89.74 & 70.83 & 90.00 & 85.00 & \textbf{79.69} \\
\textbf{(Ours)} & Kurtosis & 50 : 50 & 75.00 & 53.33 & 69.23 & 75.00 & 100.00 & 80.00 & 73.44 \\
& CV & 50 : 50 & 55.00 & 73.33 & 82.05 & 70.83 & 80.00 & 65.00 & 71.88 \\
\bottomrule
\end{tabular*}
\caption{
    Detailed Performance on ALFWorld (Qwen3-8B). 
    Success Rate (\%) across six task types on (a) Seen and (b) Unseen splits. SFT:RL indicates the allocation ratio. PRISM outperforms all hybrid baselines while requiring only half the data of the sequential SFT-then-RL pipeline.
    }
\label{tab:alfworld}
\end{table*}

\begin{table}[t]
    \centering
    \small % 空间释放后，可以尝试用更大的 small 字体
    \renewcommand{\arraystretch}{1.2}
    \setlength{\tabcolsep}{0pt}
    
    \begin{tabular*}{\columnwidth}{@{\extracolsep{\fill}} l cc cc cc}
    \toprule
    & \multicolumn{2}{c}{\textbf{Split (\%)}} & \multicolumn{2}{c}{\textbf{Qwen3-8B}} & \multicolumn{2}{c}{\textbf{Llama-3.1-8B}} \\
    \cmidrule{2-3} \cmidrule{4-5} \cmidrule{6-7}
    \textbf{Method} & \textbf{SFT} & \textbf{RL} & \textbf{SR(\%)} & \textbf{Score} & \textbf{SR(\%)} & \textbf{Score} \\
    \midrule
    Base Model & - & - & 17.19 & 28.78 & 1.56 & 1.56 \\
    \midrule
    SFT & 100 & - & 42.97 & 73.87 & 18.75 & 29.34 \\
    GRPO & - & 100 & 46.88 & 68.80 & 51.56 & 70.66 \\
    GiGPO & - & 100 & 46.09 & 71.42 & 52.34 & 73.81 \\
    \midrule
    Random & 50 & 50 & 55.47 & 78.21 & 54.69 & 79.40 \\
    HPT & 50 & 50 & 54.69 & 75.48 & 55.47 & 80.46 \\
    SFT-then-RL & 100 & 100 & 59.38 & 80.82 & 60.16 & 81.65 \\
    \midrule[0.8pt]
    PRISM (Gini) & 50 & 50 & \textbf{64.84} & \textbf{85.15} & \textbf{68.75} & \textbf{84.82} \\
    PRISM (Kurt) & 50 & 50 & 63.28 & 83.87 & 64.06 & 81.79 \\
    PRISM (CV) & 50 & 50 & 61.74 & 84.33 & 61.72 & 81.55 \\
    \bottomrule
    \end{tabular*}
\caption{
Main Results on WebShop. We compare PRISM against baselines using different data allocation strategies. Data Split denotes the number of trajectories utilized for SFT and RL phases respectively. PRISM consistently outperforms the sequential baseline (SFT-then-RL) using only 50\% of the total training budget, demonstrating the efficiency of concentration-aware data arbitration.
}
\label{tab:webshop_main}
\end{table}

\section{Experiments}
\label{sec:experiments}
\subsection{Experimental Setup}
\label{sec:exp_setup}

\paragraph{Benchmarks.}
Evaluation is conducted on two representative agentic benchmarks requiring distinct cognitive capabilities. \textbf{WebShop}~\citep{yao2022webshop}, an interactive e-commerce environment, assesses the agent's capacity for instruction following and attribute matching over long horizons, simulating real-world website navigation. Complementarily, \textbf{ALFWorld}~\citep{shridhar2020alfworld} provides a text-based embodied simulation that demands compositional generalization for household tasks. Following standard protocols, performance is reported on both \textbf{Seen} (training distribution) and \textbf{Unseen} (generalization) splits to rigorously assess robustness against environmental shifts.

\paragraph{Implementation Details.} Using \textbf{Qwen3-8B}~\citep{qwen3} and \textbf{Llama-3.1-8B-Instruct}~\citep{llama3modelcard} as backbones, we implement SFT via \textbf{Llama-Factory}~\citep{zheng2024llamafactory} and RL (GRPO) via \textbf{verl-agent}~\citep{feng2025group}. PRISM initiates with a \textbf{gradient probing phase} on the frozen base model to compute concentration metrics. These metrics serve as a filter to disentangle the dataset: low-conflict samples are assigned to SFT, while high-conflict samples are routed to RL. This selective allocation minimizes computational overhead while maximizing structural adaptation. See Appendix~\ref{sec:appendix_Implementation_Details} for full hyperparameters.

\paragraph{Baselines} We evaluate PRISM against three distinct categories:
(1) \textbf{Monolithic Baselines} (100\% budget): \textbf{SFT}, \textbf{GRPO}, and \textbf{GiGPO}~\citep{feng2025group}, the current state-of-the-art method for agentic RL;
(2) \textbf{Iso-Compute Baselines} (50/50 split): \textbf{Random} selection (serving as a control) and \textbf{HPT}~\citep{lv2025towards}, a leading outcome-aware method based on pass rates;
and (3) \textbf{Canonical Pipeline}: \textbf{SFT-then-RL} (100\%+100\%), which serves as a compute-intensive upper bound.
Crucially, to ensure robustness, all reported results represent the mean across three random seeds.

\subsection{Main Results}
\label{sec:main_results}

\paragraph{Generalization and Efficiency on ALFWorld}
As detailed in Table \ref{tab:alfworld}, PRISM demonstrates superior task mastery and generalization capabilities. On \textbf{Seen} tasks (in-distribution), PRISM (Gini) achieves a remarkable success rate of \textbf{95.31\%}, significantly outperforming the sequential SFT-then-RL baseline (88.28\%) and standard GRPO (67.19\%). 
Crucially, on \textbf{Unseen} tasks (out-of-distribution), PRISM exhibits exceptional robustness, reaching a success rate of \textbf{79.69\%}. This represents a substantial \textbf{10.16\% absolute improvement} over the sequential baseline. 

These results validate our core hypothesis: routing low-conflict data to SFT facilitates efficient pattern consolidation, while reserving high-conflict data for RL drives critical structural adaptation. Unlike standard RL, which risks overfitting to environmental noise when trained on full data, PRISM selectively targets trajectories requiring logical restructuring. Consequently, it achieves these gains using only \textbf{50\% of the RL compute budget}, effectively mitigating optimization interference and preserving the model's structural plasticity for novel scenarios.

\paragraph{Backbone-Agnostic Robustness on WebShop}
Table \ref{tab:webshop_main} highlights PRISM's decisive advantage in interactive decision-making across diverse model architectures. 
On \textbf{Qwen3-8B}, PRISM (Gini) establishes a new state-of-the-art with a score of \textbf{85.15} and a success rate of \textbf{64.84\%}, surpassing both the outcome-aware baseline HPT (75.48) and the compute-intensive SFT-then-RL (80.82). 
Notably, this superiority extends to \textbf{Llama-3.1-8B}, where PRISM improves the Success Rate by \textbf{+8.59\%} over the sequential baseline (68.75\% vs. 60.16\%). 

The consistent performance of the Gini metric across both benchmarks suggests that \textbf{spatial gradient concentration} serves as a robust proxy for cognitive dissonance in agentic tasks. By filtering out diffuse, template-like samples for SFT, PRISM ensures that the expensive RL phase is exclusively dedicated to resolving high-conflict bottlenecks (e.g., complex attribute matching), thereby preventing the gradient dilution often observed in indiscriminate full-data training.

\begin{table*}[t]
    \centering
    \small
    \renewcommand{\arraystretch}{1.3} 
    \setlength{\tabcolsep}{0pt}
    
    \begin{tabular*}{\textwidth}{@{\extracolsep{\fill}} l l c @{\hskip 15pt} cccc c}
    \toprule
    & & \textbf{Data Allocation} & \multicolumn{4}{c}{\textbf{Training Wall-clock Time}} & \\
    \cmidrule{4-7}
    \textbf{Task} & \textbf{Method} & \textbf{(SFT : RL)} & \textbf{Probing} & \textbf{SFT Phase} & \textbf{RL Phase} & \textbf{Total Time} & \textbf{Speedup} \\
    \midrule
    
    \multirow{6}{*}{\textbf{WebShop}} 
    & GRPO   & 0\% : 100\% & -      & -      & 5h 53m & 5h 53m & $1.00\times$ \\
    & Random (50\%)     & 50\% : 50\% & -      & 8m     & 3h 16m & 3h 24m & $1.73\times$ \\
    \cmidrule{2-8}
    & PRISM (CV)       & 50\% : 50\% & 1m 48s & 8m     & 3h 09m & 3h 18m & $1.77\times$ \\
    & PRISM (Gini)     & 50\% : 50\% & 1m 48s & 8m     & 3h 10m & 3h 19m & $1.76\times$ \\
    & PRISM (Kurtosis) & 50\% : 50\% & 1m 48s & 8m     & 2h 51m & \textbf{3h 00m} & \textbf{1.95$\times$} \\
    
    \midrule[1pt]
    
    \multirow{6}{*}{\textbf{ ALFWorld}} 
    & GRPO  & 0\% : 100\% & -      & -      & 36h 13m & 36h 13m & $1.00\times$ \\
    & Random (50\%)     & 50\% : 50\% & -      & 7m     & 11h 41m & 11h 48m & $3.07\times$ \\
    \cmidrule{2-8}
    & PRISM (CV)       & 50\% : 50\% & 2m 16s & 7m     & 11h 26m & 11h 35m & $3.12\times$ \\
    & PRISM (Gini)     & 50\% : 50\% & 2m 16s & 7m     & 11h 06m & \textbf{11h 15m} & \textbf{3.22$\times$} \\
    & PRISM (Kurtosis) & 50\% : 50\% & 2m 16s & 7m     & 11h 11m & 11h 20m & $3.20\times$ \\
    
    \bottomrule
    \end{tabular*}

\caption{
    Computational Efficiency and Training Costs. 
    Wall-clock time comparison on 8$\times$ NVIDIA A100 (80GB) GPUs. 
    Data Allocation specifies the proportion of the dataset assigned to the SFT and RL phases, respectively. 
   PRISM achieves superior results by intelligently partitioning a single dataset into optimal learning regimes, yielding a maximum speedup of \textbf{3.22$\times$}.
}
    \label{tab:raw_time_stats}
\end{table*}

\section{Ablation and Analysis}
\subsection{Validating the Logic of Conflict-Aware Routing}
\begin{figure}[t]
    \centering
    % Placeholder for your preliminary plot
    \includegraphics[width=\linewidth]{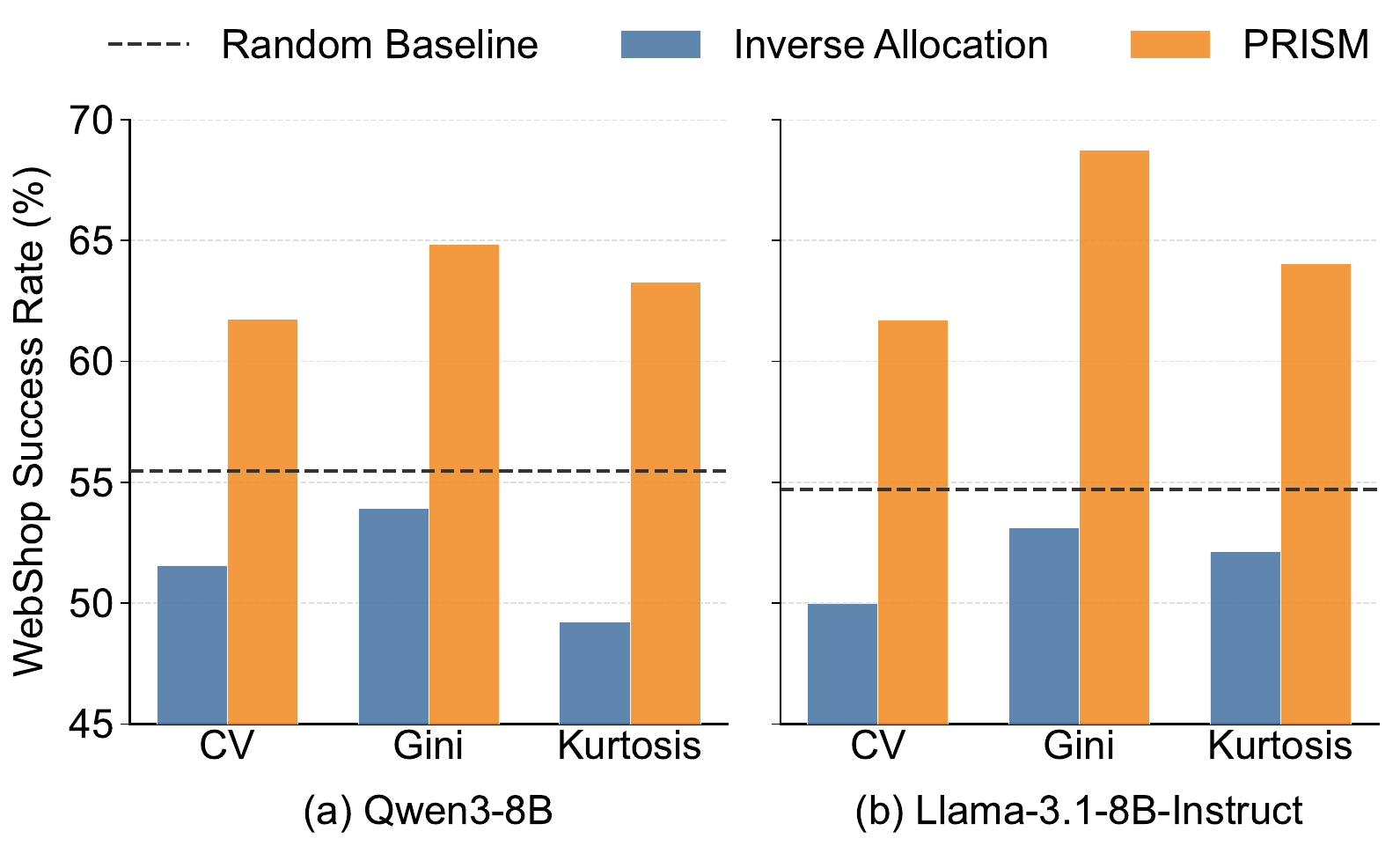}
    \caption{Ablation study of data routing strategies on WebShop. We compare PRISM (orange) with the Inverse allocation (blue: SFT on high-conflict data and RL on low-conflict data) under three concentration metrics for (a) Qwen3-8B and (b) Llama-3.1-8B-Instruct. The dashed line denotes the Random Baseline.}
    \label{fig:prelim_analysis}
\end{figure}

To validate the causal link, we partitioned data into high- and low-concentration subsets using the Gini, Kurtosis, and CV. We compared applying RL to high-concentration and SFT to low-concentration data against the reverse configuration (SFT on high/RL on low). As shown in Figure \ref{fig:prelim_analysis}, prioritizing RL for high-concentration data significantly outperforms both the random baseline and the reverse setup. This confirms that concentrated updates signal structural conflicts necessitating exploration, whereas forcing RL on low-conflict data disrupts consolidated norms~\citep{chizat2019lazy}. Thus, selective allocation based on concentration metrics is empirically superior.

\renewcommand{\tabularxcolumn}[1]{m{#1}}

\newcolumntype{Y}{>{\centering\arraybackslash}X}

\begin{table}[t]
    \centering
    \small
    \renewcommand{\arraystretch}{1.5} 
    \begin{tabularx}{0.95\linewidth}{c Y Y} 
        \toprule
        \textbf{Routing Metric} & \textbf{WebShop} (Score) & \textbf{ALFWorld} (SR \%) \\
        \midrule
        Gradient Magnitude ($L_2$) & 79.75 & 90.63 \\
        \textbf{PRISM (Spatial Gini)} & \textbf{85.15} & \textbf{95.31} \\
        \bottomrule
    \end{tabularx}
    \caption{\textbf{Spatial Concentration vs. Gradient Magnitude.} We compare magnitude-based routing (allocating the top 50\% of samples by magnitude to RL) against PRISM. Results show spatial concentration identifies structural adaptation requirements missed by gradient magnitude alone.}
    \label{tab:ablation_loss}
\end{table}

\subsection{Disentangling Structural Conflict from Update Intensity}
\label{sec:ablation_magnitude_vs_intensity}

A critical question is whether PRISM simply proxies sample difficulty. Comparing PRISM against a Gradient Magnitude ( routing top 50\% samples by $L_2$ norm to RL) in Table \ref{tab:ablation_loss} reveals a decisive advantage (+5.4\% on WebShop). This distinction is grounded in optimization dynamics: \textbf{High Magnitude $\neq$ RL Need.} Large gradient norms often indicate simple ``knowledge gaps'' (e.g., unfamiliar entities) that are structurally compatible with the model, making them ideal for efficient pattern consolidation via SFT rather than expensive RL exploration~\citep{paul2021deep}. Magnitude-based routing inefficiently misallocates these learnable samples to RL. In contrast, high concentration signals structural conflict. Concentrated updates imply that the necessary logic correction is localized within specific functional units (e.g., Knowledge Neurons), reflecting a fundamental inconsistency that requires the exploratory adaptation of RL to resolve~\citep{dai2022knowledge, simsekli2019tail}.

\subsection{Sensitivity to Allocation Ratio}
\label{sec:ablation_ratio}

\begin{figure}[h]
\centering
\small
\includegraphics[width=0.85\columnwidth]{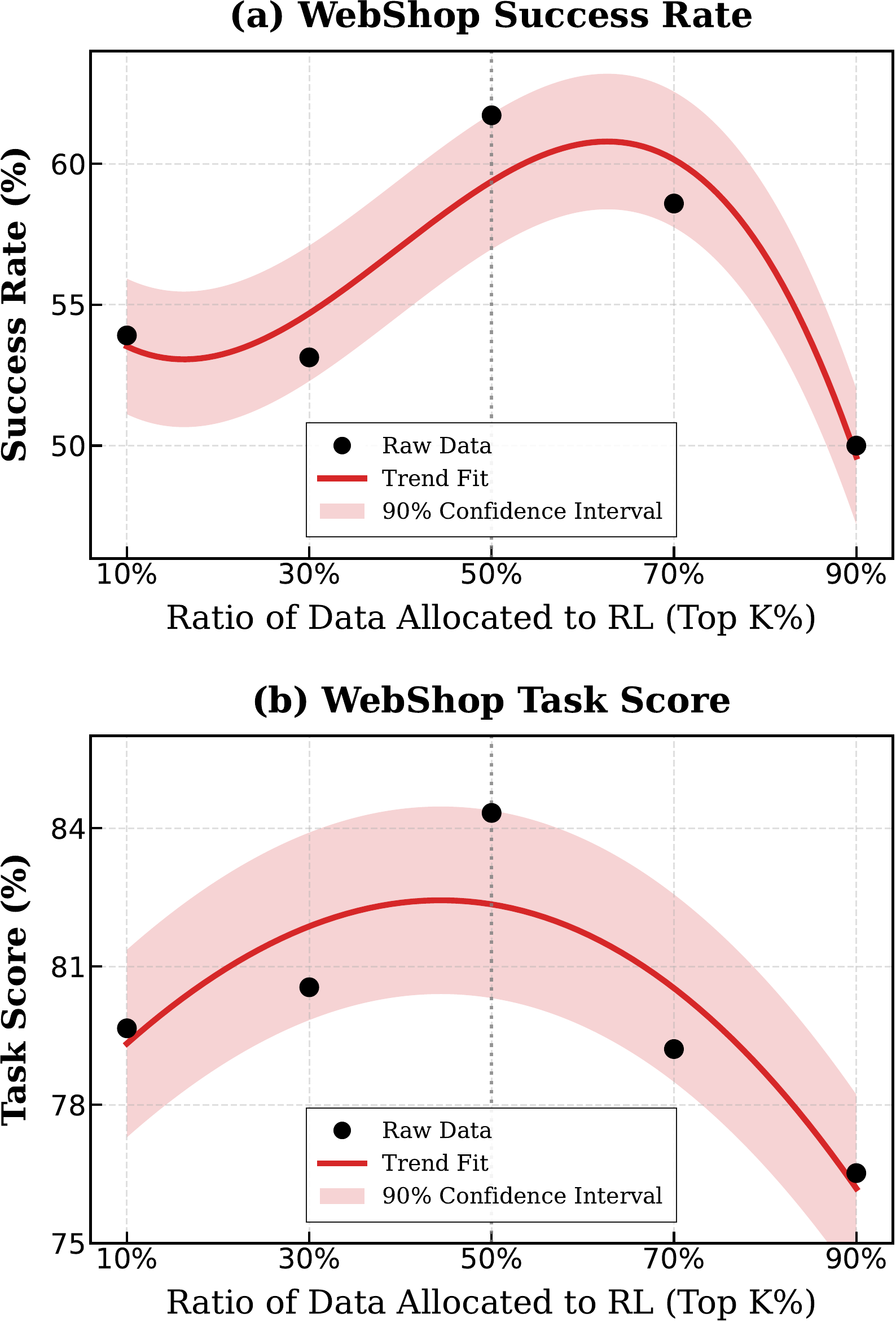}
\caption{\textbf{Sensitivity to RL Allocation Ratio.} 
Performance of Qwen3-8B on WebShop (CV metric) across varying RL data proportions. 
The observed inverted U-shape peaks at a 50\% split, indicating that a balanced allocation yields optimal performance compared to insufficient adaptation or excessive exploration.}
\label{fig:ratio_ablation}
\end{figure}

We evaluate PRISM's sensitivity to routing thresholds by varying the RL allocation ratio. As shown in Figure~\ref{fig:ratio_ablation}, performance exhibits a distinct inverted U-shape peaking near 50\%. This revealing trend highlights a critical trade-off: insufficient RL allocation ($<30\%$) provides inadequate structural adaptation for high-conflict samples, while excessive allocation ($>70\%$) leads to gradient dilution. Specifically, forcing RL on low-conflict data injects exploratory noise into trivial behaviors, thereby contaminating the gradients and interfering with previously consolidated patterns.

\subsection{Pareto Improvement: Optimization Disentanglement}
\label{sec:efficiency_analysis}
Beyond raw performance, PRISM fundamentally optimizes the computational utility of agent training. As detailed in Table \ref{tab:raw_time_stats}, our framework reduces wall-clock time by \textbf{nearly 50\%} on WebShop and achieves a \textbf{3.22$\times$ speedup} on ALFWorld compared to full-data RL.
More importantly, contrasting our results with the SFT-then-RL baseline reveals a critical insight regarding data scaling. While the sequential baseline processes 100\% of the data via computationally expensive RL to achieve competitive results, PRISM achieves superior performance using only $\sim$50\% of the RL budget. This indicates that blindly forcing exploration on well-consolidated knowledge yields diminishing returns.

PRISM effectively disentangles the optimization objectives: it delegates pattern consolidation to the cost-efficient SFT phase, while reserving the expensive RL budget for trajectories requiring structural restructuring of key neurons. This synergy realizes a Pareto improvement in the performance-efficiency trade-off, proving that smarter data arbitration is superior to exhaustive exploration.

\section{Related Works}

\paragraph{Data Allocation in SFT--RL.} Data allocation balances imitation and exploration, yet current paradigms often rely on coarse heuristics that ignore the interplay between model state and data difficulty. Monolithic sequencing~\citep{ouyang2022training} uses rigid schedules, failing to distinguish between pattern consolidation and structural adaptation~\citep{zhou2023lima}. Similarly, universal exploration~\citep{shao2024deepseekmath, feng2025group} indiscriminately applies RL, which can trigger optimization instability on high-conflict data lacking SFT-consolidated priors~\citep{deepseek2025r1}. While outcome-centric filtering~\citep{lv2025towards} uses external proxies, it suffers from an ``observational gap'' where correctness masks latent shortcuts or unfaithful reasoning~\citep{geirhos2020shortcut, dziri2023faith}. In contrast, PRISM shifts to internal learning dynamics, utilizing the spatial geometric structure of gradients to quantify intrinsic conflict. This enables granular arbitration based on actual cognitive demand rather than rigid pipeline orders or outcome-based heuristics.

\paragraph{Gradient-Based Diagnostics} Gradients provide high-fidelity diagnostics for internal dynamics and functional specialization~\citep{zhao2025beyond, dai2022knowledge}. Unlike intensity- or similarity-based approaches~\citep{paul2021deep, zhao2024deciphering}, PRISM leverages the spatial geometric structure of updates, aligning with mechanistic evidence of localized representations~\citep{geva2021transformer}. Specifically, concentrated updates signal structural restructuring to resolve logical inconsistencies~\citep{simsekli2019tail, meng2022locating}, while diffuse updates reflect knowledge compatibility and consolidation~\citep{chizat2019lazy, agarwal2022estimating}. We repurpose these signals into a proactive arbitration mechanism for optimal regime routing.

\section{Conclusion}
\label{sec:conclusion}
In this work, we introduced \textbf{PRISM}, a framework that bridges cognitive learning principles with neural optimization to resolve the long-standing data arbitration challenge in agent training. By utilizing the spatial geometric structure of gradients as an intrinsic diagnostic for cognitive conflict, PRISM effectively disentangles the training process into pattern consolidation via SFT and structural adaptation via RL. Our results confirm that precision in data routing outweighs raw volume: PRISM not only establishes new state-of-the-art benchmarks but also mitigates optimization interference, yielding superior generalization. This approach represents a significant \textbf{Pareto improvement}, achieving these gains with a \textbf{3.22$\times$} training speedup. Ultimately, PRISM marks a shift from heuristic-based pipelines toward a principled, dynamics-aware orchestration of LLM post-training.

\section{Limitations}
Despite its robust performance and efficiency, PRISM has several limitations that warrant further exploration. First, due to \textbf{computational constraints}, our evaluation is primarily focused on 7B–8B parameter models. While we hypothesize that the \textbf{spatial geometric structure} of gradients is a scale-invariant mechanistic property of Transformers, extensive verification on large-scale models (e.g., 70B+ parameters) remains for future work. Second, we currently employ a \textbf{static routing strategy} based on initial gradient concentration to isolate diagnostic signals and minimize computational overhead. This approach does not account for the dynamic evolution of a model's internal state, where a high-conflict sample might transition into a routine consolidation candidate as training progresses. Finally, our scope is concentrated on \textbf{agentic decision-making} benchmarks. While these tasks effectively highlight the functional divergence between SFT and RL, the generalizability of our gradient-based diagnostic to other complex domains, such as advanced mathematical reasoning or open-ended creative generation, requires further empirical investigation.

\section{Acknowledgements}
The research in this article is supported by the New Generation Artificial Intelligence of China (2024YFE0203700), National Natural Science Foundation of China under Grants U22B2059 and 62576124.

\bibliography{custom}

@article{qian2025toolrl,
  title={Toolrl: Reward is all tool learning needs},
  author={Qian, Cheng and Acikgoz, Emre Can and He, Qi and Wang, Hongru and Chen, Xiusi and Hakkani-T{\"u}r, Dilek and Tur, Gokhan and Ji, Heng},
  journal={arXiv preprint arXiv:2504.13958},
  year={2025}
}

@article{lv2025towards,
  title={Towards a unified view of large language model post-training},
  author={Lv, Xingtai and Zuo, Yuxin and Sun, Youbang and Liu, Hongyi and Wei, Yuntian and Chen, Zhekai and Zhu, Xuekai and Zhang, Kaiyan and Wang, Bingning and Ding, Ning and others},
  journal={arXiv preprint arXiv:2509.04419},
  year={2025}
}

@article{zhou2023lima,
  title={Lima: Less is more for alignment},
  author={Zhou, Chunting and Liu, Pengfei and Xu, Puxin and Iyer, Srinivasan and Sun, Jiao and Mao, Yuning and Ma, Xuezhe and Efrat, Avia and Yu, Ping and Yu, Lili and others},
  journal={Advances in Neural Information Processing Systems},
  volume={36},
  pages={55006--55021},
  year={2023}
}

@article{paul2021deep,
  title={Deep learning on a data diet: Finding important examples early in training},
  author={Paul, Mansheej and Ganguli, Surya and Dziugaite, Gintare Karolina},
  journal={Advances in neural information processing systems},
  volume={34},
  pages={20596--20607},
  year={2021}
}

@inproceedings{agarwal2022estimating,
  title={Estimating example difficulty using variance of gradients},
  author={Agarwal, Chirag and D'souza, Daniel and Hooker, Sara},
  booktitle={Proceedings of the IEEE/CVF Conference on Computer Vision and Pattern Recognition},
  pages={10368--10378},
  year={2022}
}

@article{chizat2019lazy,
  title={On lazy training in differentiable programming},
  author={Chizat, Lenaic and Oyallon, Edouard and Bach, Francis},
  journal={Advances in neural information processing systems},
  volume={32},
  year={2019}
}

@inproceedings{simsekli2019tail,
  title={A tail-index analysis of stochastic gradient noise in deep neural networks},
  author={Simsekli, Umut and Sagun, Levent and Gurbuzbalaban, Mert},
  booktitle={International Conference on Machine Learning},
  pages={5827--5837},
  year={2019},
  organization={PMLR}
}

@article{yao2022webshop,
  title={Webshop: Towards scalable real-world web interaction with grounded language agents},
  author={Yao, Shunyu and Chen, Howard and Yang, John and Narasimhan, Karthik},
  journal={Advances in Neural Information Processing Systems},
  volume={35},
  pages={20744--20757},
  year={2022}
}

@article{feng2025group,
  title={Group-in-group policy optimization for llm agent training},
  author={Feng, Lang and Xue, Zhenghai and Liu, Tingcong and An, Bo},
  journal={arXiv preprint arXiv:2505.10978},
  year={2025}
}

@article{zhang2025generalizability,
  title={Generalizability of Large Language Model-Based Agents: A Comprehensive Survey},
  author={Zhang, Minxing and Yang, Yi and Xie, Roy and Dhingra, Bhuwan and Zhou, Shuyan and Pei, Jian},
  journal={arXiv preprint arXiv:2509.16330},
  year={2025}
}

@inproceedings{qin2024toolllm,
  title={ToolLLM: Facilitating Large Language Models to Master 16000+ Real-world APIs},
  author={Qin, Yujia and Liang, Shihao and Ye, Yining and Zhu, Kunlun and Yan, Lan and Lu, Yaxi and Lin, Yankai and Cong, Xin and Tang, Xiangru and Qian, Bill and others},
  booktitle={ICLR},
  year={2024}
}

@article{chu2025sft,
  title={Sft memorizes, rl generalizes: A comparative study of foundation model post-training},
  author={Chu, Tianzhe and Zhai, Yuexiang and Yang, Jihan and Tong, Shengbang and Xie, Saining and Schuurmans, Dale and Le, Quoc V and Levine, Sergey and Ma, Yi},
  journal={arXiv preprint arXiv:2501.17161},
  year={2025}
}

@article{shao2024deepseekmath,
  title={Deepseekmath: Pushing the limits of mathematical reasoning in open language models},
  author={Shao, Zhihong and Wang, Peiyi and Zhu, Qihao and Xu, Runxin and Song, Junxiao and Bi, Xiao and Zhang, Haowei and Zhang, Mingchuan and Li, YK and Wu, Yang and others},
  journal={arXiv preprint arXiv:2402.03300},
  year={2024}
}

@article{ouyang2022training,
  title={Training language models to follow instructions with human feedback},
  author={Ouyang, Long and Wu, Jeffrey and Jiang, Xu and Almeida, Diogo and Wainwright, Carroll and Mishkin, Pamela and Zhang, Chong and Agarwal, Sandhini and Slama, Katarina and Ray, Alex and others},
  journal={Advances in neural information processing systems},
  volume={35},
  pages={27730--27744},
  year={2022}
}

@article{llama3modelcard,
  title={The llama 3 herd of models},
  author={Grattafiori, Aaron and Dubey, Abhimanyu and Jauhri, Abhinav and Pandey, Abhinav and Kadian, Abhishek and Al-Dahle, Ahmad and Letman, Aiesha and Mathur, Akhil and Schelten, Alan and Vaughan, Alex and others},
  journal={arXiv preprint arXiv:2407.21783},
  year={2024}
}

@book{piaget1952origins,
  title={The origins of intelligence in children},
  author={Piaget, Jean},
  volume={8},
  year={1952},
  publisher={International Universities Press New York}
}

@inproceedings{shridhar2020alfworld,
  title={ALFWorld: Aligning Text and Embodied Environments for Interactive Learning},
  author={Shridhar, Mohit and Yuan, Xingdi and C{\^o}t{\'e}, Marc-Alexandre and Bisk, Yonatan and Trischler, Adam and Hausknecht, Matthew},
  booktitle={ICLR},
  year={2021}
}

@inproceedings{zhao2024deciphering,
  title={Deciphering the impact of pretraining data on large language models through machine unlearning},
  author={Zhao, Yang and Du, Li and Ding, Xiao and Xiong, Kai and Sun, Zhouhao and Jun, Shi and Liu, Ting and Qin, Bing},
  booktitle={Findings of the Association for Computational Linguistics: ACL 2024},
  pages={9386--9406},
  year={2024}
}

@inproceedings{zhao2025beyond,
  title={Beyond similarity: A gradient-based graph method for instruction tuning data selection},
  author={Zhao, Yang and Du, Li and Ding, Xiao and Ouyang, Yangou and Wang, Hepeng and Xiong, Kai and Gao, Jinglong and Sun, Zhouhao and Xu, Dongliang and Yang, Qing and others},
  booktitle={Proceedings of the 63rd Annual Meeting of the Association for Computational Linguistics (Volume 1: Long Papers)},
  pages={24391--24404},
  year={2025}
}

@article{qwen3,
    title={Qwen3 Technical Report}, 
    author={An Yang and Anfeng Li and Baosong Yang and Beichen Zhang and Binyuan Hui and Bo Zheng and Bowen Yu and Chang Gao and Chengen Huang and Chenxu Lv and Chujie Zheng and Dayiheng Liu and Fan Zhou and Fei Huang and Feng Hu and Hao Ge and Haoran Wei and Huan Lin and Jialong Tang and Jian Yang and Jianhong Tu and Jianwei Zhang and Jianxin Yang and Jiaxi Yang and Jing Zhou and Jingren Zhou and Junyang Lin and Kai Dang and Keqin Bao and Kexin Yang and Le Yu and Lianghao Deng and Mei Li and Mingfeng Xue and Mingze Li and Pei Zhang and Peng Wang and Qin Zhu and Rui Men and Ruize Gao and Shixuan Liu and Shuang Luo and Tianhao Li and Tianyi Tang and Wenbiao Yin and Xingzhang Ren and Xinyu Wang and Xinyu Zhang and Xuancheng Ren and Yang Fan and Yang Su and Yichang Zhang and Yinger Zhang and Yu Wan and Yuqiong Liu and Zekun Wang and Zeyu Cui and Zhenru Zhang and Zhipeng Zhou and Zihan Qiu},
    journal = {arXiv preprint arXiv:2505.09388},
    year={2025}
}

@inproceedings{zheng2024llamafactory,
  title={Llamafactory: Unified efficient fine-tuning of 100+ language models},
  author={Zheng, Yaowei and Zhang, Richong and Zhang, Junhao and Ye, Yanhan and Luo, Zheyan},
  booktitle={Proceedings of the 62nd annual meeting of the association for computational linguistics (volume 3: system demonstrations)},
  pages={400--410},
  year={2024}
}

@inproceedings{dai2022knowledge,
  title={Knowledge neurons in pretrained transformers},
  author={Dai, Damai and Dong, Li and Hao, Yaru and Sui, Zhifang and Chang, Baobao and Wei, Furu},
  booktitle={Proceedings of the 60th Annual Meeting of the Association for Computational Linguistics (Volume 1: Long Papers)},
  pages={8493--8502},
  year={2022}
}

@article{meng2022locating,
  title={Locating and editing factual associations in gpt},
  author={Meng, Kevin and Bau, David and Andonian, Alex and Belinkov, Yonatan},
  journal={Advances in neural information processing systems},
  volume={35},
  pages={17359--17372},
  year={2022}
}

@article{deepseek2025r1,
  title={DeepSeek-R1: Incentivizing Reasoning Capability in LLMs via Reinforcement Learning},
  author={DeepSeek-AI and others},
  journal={arXiv preprint arXiv:2501.12948},
  year={2025}
}

@article{geirhos2020shortcut,
  title={Shortcut learning in deep neural networks},
  author={Geirhos, Robert and others},
  journal={Nature Machine Intelligence},
  year={2020}
}

@article{dziri2023faith,
  title={Faith and Fate: Limits of Transformers on Compositionality},
  author={Dziri, Nouha and others},
  journal={arXiv preprint arXiv:2305.18654},
  year={2023}
}

@inproceedings{geva2021transformer,
  title={Transformer feed-forward layers are key-value memories},
  author={Geva, Mor and Schuster, Roei and Berant, Jonathan and Levy, Omer},
  booktitle={Proceedings of the 2021 Conference on Empirical Methods in Natural Language Processing},
  pages={5484--5495},
  year={2021}
}

@article{guo2025deepseek,
  title={Deepseek-r1: Incentivizing reasoning capability in llms via reinforcement learning},
  author={Guo, Daya and Yang, Dejian and Zhang, Haowei and Song, Junxiao and Zhang, Ruoyu and Xu, Runxin and Zhu, Qihao and Ma, Shirong and Wang, Peiyi and Bi, Xiao and others},
  journal={arXiv preprint arXiv:2501.12948},
  year={2025}
}

@article{schulman2017proximal,
  title={Proximal policy optimization algorithms},
  author={Schulman, John and Wolski, Filip and Dhariwal, Prafulla and Radford, Alec and Klimov, Oleg},
  journal={arXiv preprint arXiv:1707.06347},
  year={2017}
}

\appendix

\section{Details of Concentration Metrics}
\label{sec:appendix_metrics}

In this section, we provide the formal definitions for the gradient concentration metrics used to quantify cognitive dissonance.

\paragraph{Gradient Vector Construction.}
For a given trajectory $\tau_i$, let $\mathcal{L}(\tau_i)$ denote the standard next-token prediction loss, averaged over all valid tokens in the sequence. To characterize the spatial geometric structure of the model's internal response, we analyze the gradients with respect to the specific linear projection weights of the Transformer backbone. For a model with $L$ layers, we define the parameter groups for the $l$-th layer as $\mathcal{P}_l = \{W_q, W_k, W_v, W_o, W_{\text{gate}}, W_{\text{up}}, W_{\text{down}}\}$. Aggregating across all layers, we obtain a total of $N = 7L$ parameter groups. This multi-layered grouping allows us to capture the distribution of optimization effort across the network's functional units, providing the necessary resolution to measure spatial concentration.

We define the gradient concentration vector $\mathbf{g}_i \in \mathbf{R}_{\ge 0}^N$ as the collection of Frobenius norms for each parameter group's gradient matrix:
\begin{equation}
    \mathbf{g}_i = \left[ \| \nabla_{\theta_1} \mathcal{L}(\tau_i) \|_F, \dots, \| \nabla_{\theta_N} \mathcal{L}(\tau_i) \|_F \right]^\top.
\end{equation}
Let $\mu_i$ and $\sigma_i$ denote the arithmetic mean and standard deviation of the elements in $\mathbf{g}_i$, respectively. $\epsilon$ is a small constant ($1e^{-8}$) added for numerical stability.

\paragraph{1. Gini Coefficient.}
The Gini coefficient measures the inequality of the gradient contribution distribution. We first sort the elements of $\mathbf{g}_i$ in \textbf{non-decreasing order}, such that $g_{i,(1)} \le g_{i,(2)} \le \dots \le g_{i,(N)}$. The metric is calculated as:
\begin{equation}
    s_i^{\text{Gini}} = \frac{\sum_{j=1}^{N} (2j - N - 1)\, g_{i,(j)}}{N \sum_{j=1}^{N} g_{i,(j)} + \epsilon}.
\end{equation}
A higher Gini coefficient indicates that a small subset of parameter groups dominates the gradient updates (sparse activation), suggesting structural conflict.

\paragraph{2. Kurtosis.}
We employ the Fourth Standardized Moment (Pearson's Kurtosis) to quantify the ``tailedness'' of the gradient distribution. This serves as a detector for extreme outliers in optimization pressure. Given the large number of parameter groups ($N \gg 100$), we utilize the population formula without small-sample bias correction:
\begin{equation}
    s_i^{\text{Kurt}} = \frac{1}{N} \sum_{j=1}^{N} \left( \frac{g_{i,j} - \mu_i}{\sigma_i + \epsilon} \right)^{4}-3.
\end{equation}
High kurtosis implies that the gradients are characterized by infrequent but extreme updates, distinguishing "spiky" structural adaptation signals from Gaussian noise.

\paragraph{3. Coefficient of Variation (CV).}
The Coefficient of Variation provides a normalized measure of concentration, describing the extent of variability in relation to the mean of the population:
\begin{equation}
    s_i^{\text{CV}} = \frac{\sigma_i}{\mu_i + \epsilon}.
\end{equation}
This metric captures the relative instability of the update signal, serving as a robust proxy for global model dissonance.
\section{Qualitative Analysis of Routed Trajectories}
\label{sec:appendix_qualitative}
To validate the cognitive dissonance hypothesis, we manually inspected trajectories routed to distinct phases.
\begin{itemize}
    \item \textbf{SFT-Routed (Low Concentration):} Typically involve straightforward instruction following or keyword matching (e.g., ``Click the 'Search' button''). The model's priors are sufficient, resulting in diffuse gradients.
    \item \textbf{RL-Routed (High Concentration):} Involve counter-intuitive reasoning or correcting a previous error (e.g., ALFWorld: ``The apple is not in the fridge, checking the cabinet''). These induce concentrated updates as specific attention heads must be re-weighted to shift the search strategy.
\end{itemize}

\section{Implementation Details}
\label{sec:appendix_Implementation_Details}

\paragraph{Gradient Probing Configuration}
To ensure consistency between the diagnostic and training phases, the Non-Invasive Gradient Probing (Stage I) utilizes the same context length constraints as the subsequent RL stage. Specifically, input sequences are standardized to a fixed length of 2048 tokens for  ALFWorld and 4096 tokens for WebShop. Sequences exceeding these limits are truncated, while shorter ones are padded with strict masking applied during gradient computation to avoid padding bias.

\paragraph{SFT}
We implement the SFT stage using the \texttt{LLaMA-Factory} framework. We perform full-parameter fine-tuning on Qwen3-8B for 3 epochs using the AdamW optimizer. The learning rate is initialized at $1\times 10^{-5}$ with a cosine decay schedule and a warmup ratio of 0.1. We employ a per-device batch size of 4 with 4 gradient accumulation steps, training in bfloat16 precision.

\paragraph{RL}
For our method, we employ the GRPO algorithm without KL divergence penalties and set the rollout size to 8. We adopt the environment configurations and reward structures from the GiGPO framework. Specifically, the actor learning rate is set to $1\times 10^{-6}$. We use a rule-based reward function: +10 for success, 0 for failure, and a penalty of -0.1 for invalid actions. Consistent with the probing phase, we limit prompts to \textbf{2048 tokens} for ALFWorld and \textbf{4096 tokens} for WebShop, with a maximum of 50 environment steps per episode for ALFWorld and 15 for WebShop. For the GiGPO baseline reported in our experiments, we strictly follow the original hyperparameter settings provided in~\citep{feng2025group}.

\begin{figure}[t]
    \centering
    \includegraphics[width=0.8\linewidth]{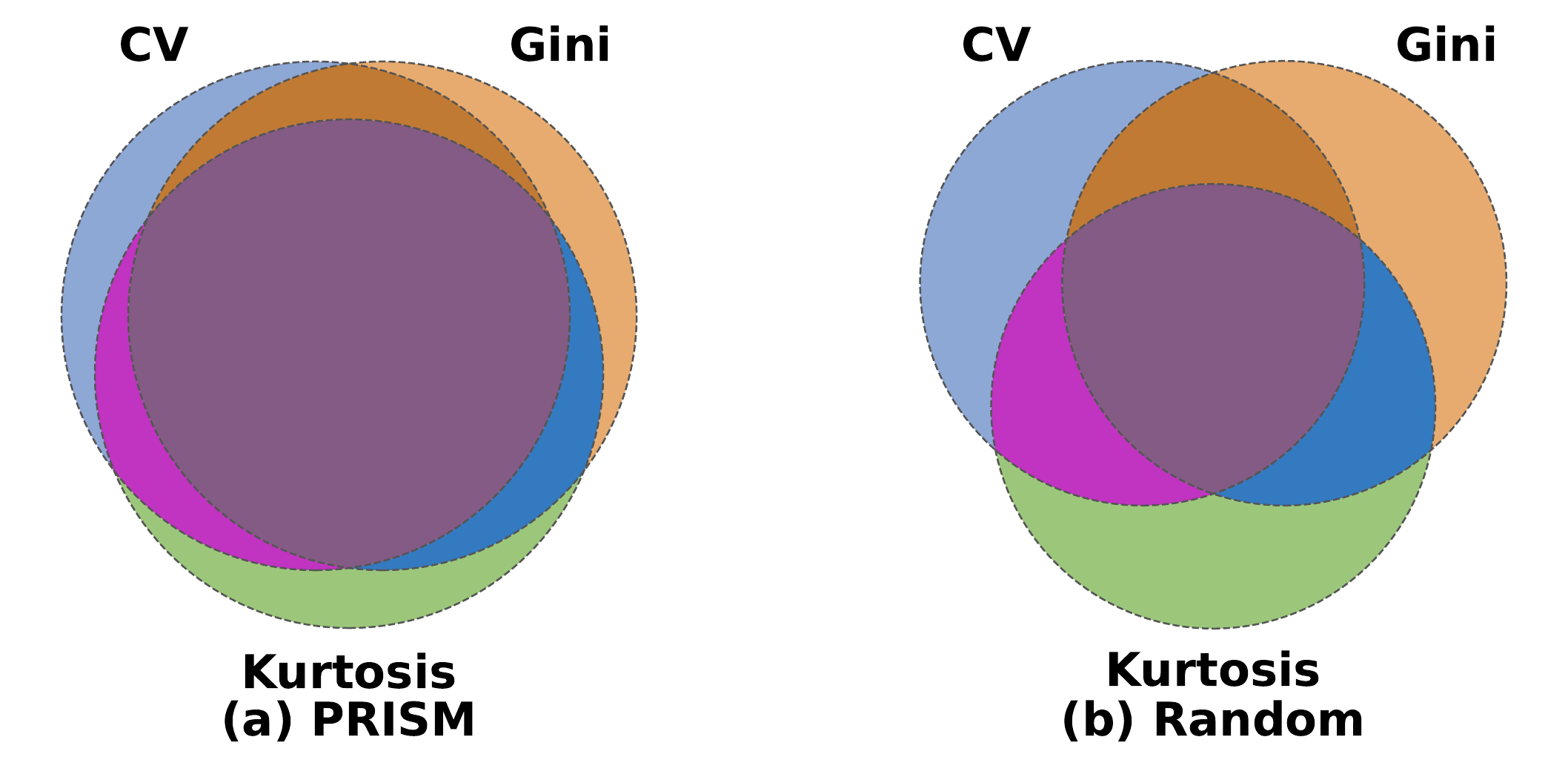}
    \caption{\textbf{Venn Diagram of Data Selection Consensus.} The intersection shows that approximately \textbf{60\%} of the high-conflict trajectories are consistently identified by all three statistical metrics. This high degree of convergence significantly exceeds the \textbf{12.5\%--25.0\%} expected from random overlapping splits, demonstrating that PRISM captures a stable underlying structural dissonance signal regardless of the specific concentration metric employed.}
    \label{fig:venn_consensus}
\end{figure}

\section{Consensus Analysis of Gradient Concentration Metrics}
\label{sec:appendix_consensus}

As illustrated in Figure \ref{fig:venn_consensus}, we observe a substantial overlap among the high-conflict subsets identified by these metrics. This empirical evidence suggests that while individual metrics may align more closely with specific task dynamics, they largely converge on a core set of high-conflict data. This consensus indicates that PRISM captures a universal and robust signal of cognitive dissonance, rather than being an artifact of specific metric selection.

\section{Experimental Environments and Task Decomposition}

We evaluate our framework on two complex agent benchmarks: \textbf{WebShop} and \textbf{ALFWorld}. These environments require the agent to demonstrate diverse capabilities, ranging from navigating e-commerce interfaces to solving interactive household tasks. 
% For supervised components that require demonstrations, we use the expert trajectory datasets of successful trajectories released by SPA-RL~\citep{wang2025spa}. Following their setup, these expert demonstrations are formatted as ReAct-style thought--action sequences and serve as the source of expert trajectories for our experiments.

\subsection{WebShop}
WebShop simulates an e-commerce website environment, requiring models to navigate interfaces to find and purchase products that match specific user attributes.

\paragraph{Evaluation Metrics.} Following the standard protocol of the WebShop benchmark~\citep{yao2022webshop}, we evaluate the performance of our agent using two primary metrics: \textbf{Average Score} and \textbf{Success Rate (SR)}.

\begin{itemize}
    \item \textbf{Average Score}: This metric measures the granularity of task completion by calculating the attribute overlap between the product purchased by the agent and the user's instruction. For each episode $i$, the environment computes a scalar score $S_i \in [0, 1]$, which is a weighted sum of rewards based on four dimensions: product category matching, attribute recall, option selection accuracy, and price constraints. Formally, the score is calculated as:
\begin{equation}
\begin{aligned}
S_i &= \text{TypeScore} \\
&\quad \times \left( \frac{N_{attr} + N_{option} + \mathbb{I}_{price}}{N_{total}} \right)
\end{aligned}
\end{equation}

    where $N_{attr}$ and $N_{option}$ denote the number of matched attributes and options respectively, and $\mathbb{I}_{price}$ is an indicator function for price satisfaction. We report the mean score averaged across all test episodes.

    \item \textbf{Success Rate (SR)}: This is a stricter metric evaluating the agent's ability to perfectly satisfy user goals. An episode is considered successful if and only if the agent achieves a perfect score (i.e., $S_i = 1.0$). This implies that the purchased item meets all specified criteria, including correct category, attributes, options, and price limits. SR denotes the percentage of episodes where the agent successfully completed the task.
\end{itemize}

\subsection{ALFWorld}
ALFWorld aligns TextWorld with the ALFRED benchmark, consisting of interactive household tasks that require multi-step reasoning and decision-making.

\paragraph{Task Decomposition.} We report results across six ALFWorld sub-task categories: Pick (single-object retrieval), Look (object search/navigation), Clean (cleaning appliances), Heat (heating state transitions), Cool (cooling state transitions), and Pick2 (two-object pick-and-place).

\paragraph{Evaluation Metrics.} Similar to WebShop, we report the \textbf{Success Rate (SR)} for ALFWorld. An episode is considered successful if the agent completes the goal state within the maximum number of steps. We report both the overall SR and the task-wise SR for the six categories mentioned above.

\section{Theoretical Motivation: Why High-Conflict Trajectories Benefit from RL Exploration}
\label{sec:appendix_theoretical_motivation}

The main text argues that high gradient concentration indicates a structural mismatch between the current policy and the target behavior, motivating routing such trajectories to RL. This appendix provides a mechanistic explanation for why exploration-based, group-relative RL (e.g., GRPO) is well-matched to this regime. We present the argument as an intuition consistent with policy-gradient learning dynamics, rather than as a formal equivalence between gradient concentration under SFT and the RL training signal.

\paragraph{1. High conflict tends to create distinct rollout modes under sampling.}
When a state-action decision is aligned with the model's current behavior, stochastic sampling from $\pi_\theta$ often produces similar trajectories with small qualitative variation. In contrast, under structural mismatch, the policy is more likely to admit multiple competing action modes for the same state (e.g., relying on superficial heuristics versus executing a faithful reasoning chain). As a result, sampling can expose distinct outcome patterns (success/failure, or different intermediate behaviors), creating the diversity necessary for trial-and-error refinement in policy optimization~\citep{schulman2017proximal}.

\paragraph{2. GRPO is most informative when there is within-group contrast.}
GRPO-style learning constructs its update from relative comparisons within a sampled group of trajectories (e.g., group-relative advantages), rather than from matching a single reference trace~\citep{shao2024deepseekmath, feng2025group}. This implies a simple requirement: the sampled group must contain meaningfully different outcomes for the relative signal to be discriminative.
\begin{itemize}
    \item \textbf{Low-conflict regime: limited contrast yields weakly discriminative relative feedback.}
    For consolidated trajectories, sampled rollouts tend to be homogeneous in outcomes and rewards. In this case, group-relative normalization/ranking provides little separation between trajectories, so the relative learning signal becomes less informative and can be sensitive to stochasticity without yielding systematic improvement~\citep{shao2024deepseekmath, feng2025group}.

    \item \textbf{High-conflict regime: outcome separation enables contrastive credit assignment.}
    Under structural mismatch, sampling is more likely to produce both better and worse rollouts with distinct reward profiles. This within-group separation makes group-relative updates informative: the optimizer can assign credit by reinforcing behaviors that lead to verified success and suppressing those leading to failure, without requiring imitation of a single fixed trace~\citep{shao2024deepseekmath, feng2025group}. 
\end{itemize}

\paragraph{3. Exploration supports selective policy shifts where imitation can be brittle.}
A key advantage of routing high-conflict trajectories to RL is that exploration allows the learner to search over alternative behaviors and update the policy selectively based on feedback, rather than forcing the model to reproduce a particular trajectory. This is consistent with observations that RL post-training can induce behavioral improvements beyond SFT-only pipelines by leveraging reward-driven feedback to shape policy updates~\citep{ouyang2022training, guo2025deepseek}. In PRISM, this motivates concentrating RL budget on trajectories that exhibit structural mismatch, while using SFT to consolidate already-compatible behaviors.

\section{Robustness Analysis: Architecture Invariance and Confound Control}
\label{sec:appendix_robustness}

A potential concern in gradient-based analysis is whether the varying sizes of parameter matrices (e.g., $W_{down}$ vs. $W_{q}$) introduce confounds in the concentration metrics. We argue that PRISM is robust to these variations due to \textbf{Architecture Invariance}.

While larger parameter matrices naturally yield larger gradient norms, this introduces a \textbf{constant systematic bias} rather than a data-dependent variable. Since the model architecture remains static, this bias affects all trajectories identically. Moreover, our chosen metrics (e.g., Gini Coefficient) are theoretically \textbf{scale-invariant}—multiplying a subset of dimensions by a constant factor preserves the relative inequality score, effectively canceling out layer-wise scaling artifacts.

To empirically validate this, we conducted a \textbf{sensitivity analysis} by normalizing the gradient norms by the square root of the parameter count ($\|\mathbf{g}\|_F / \sqrt{N_{param}}$). We observed that this normalization resulted in \textbf{nearly identical data rankings} (Spearman's $\rho > 0.99$ on both benchmarks) compared to the raw Frobenius norms. This confirms that PRISM's median-split routing is driven by genuine structural conflict rather than architectural dimensions.

\section{AI Assistance Disclosure}
\label{sec:ai_disclosure}

We acknowledge the use of AI tools solely for language polishing and grammatical editing to improve the readability of this manuscript. All scientific claims, experimental data, and empirical results were rigorously verified by the human authors to ensure authenticity and accuracy.
\end{document}